\documentclass[a4paper]{article}
\pdfoutput=1
\usepackage{INTERSPEECH2019,multirow,hyperref}

\title{Enriching Rare Word Representations in Neural Language Models\\by Embedding Matrix Augmentation}
\name{Yerbolat Khassanov$^1$, Zhiping Zeng$^2$, Van Tung Pham$^{1,2}$, Haihua Xu$^{2}$, Eng Siong Chng$^{1,2}$}
\address{$^1$School of Computer Science and Engineering, Nanyang Technological University, Singapore\\
$^2$Temasek Laboratories, Nanyang Technological University, Singapore}
\email{\{yerbolat002,zengzp,vantung001,haihuaxu,aseschng\}@ntu.edu.sg}

\begin{document}

\maketitle
\begin{abstract}
  % What are the neural language models (NLM) and what makes them strong?
  %The success of neural language models (NLM) is mainly attributed to its exceptional generalization capability achieved by simultaneously learning both probability distribution and dense representations of words.
  The neural language models (NLM) achieve strong generalization capability by learning the dense representation of words and using them to estimate probability distribution function.
  %However, learning representation of rare words, which have insufficient amount of training samples, is challenging problem which leads to unreliable probability estimates.
  However, learning the representation of rare words is a challenging problem causing the NLM to produce unreliable probability estimates.
  To address this problem, we propose a method to enrich representations of rare words in pre-trained NLM and consequently improve its probability estimation performance.
  The proposed method augments the word embedding matrices of pre-trained NLM while keeping other parameters unchanged.
  Specifically, our method updates the embedding vectors of rare words using embedding vectors of other semantically and syntactically similar words.
  %Importantly, the proposed method doesn't require additional training data and expensive post-training procedures.
  To evaluate the proposed method, we enrich the rare street names in the pre-trained NLM and use it to rescore $100$-best hypotheses output from the Singapore English speech recognition system. 
  %We evaluate the effectiveness of the proposed method on $n$-best hypotheses rescoring task for Singapore English speech recognition containing rare street names as a case study.
  The enriched NLM reduces the word error rate by $6\%$ relative and improves the recognition accuracy of the rare words by $16\%$ absolute as compared to the baseline NLM.
  %and achieve $6\%$ and $16\%$ relative word error rate improvements over the strong baseline NLM and Kneser-Ney smoothed 4-gram models, respectively. 

\ifx
%The neural language models (NLM) became indispensable component of speech recognition pipeline where it's used to rescore the $n$-best output hypotheses.
The success of neural language models (NLM) is mainly attributed to its exceptional generalization capability achieved by simultaneously learning both probability distribution and dense representations of words.
%Its success is mainly attributed to its exceptional generalization capability achieved by simultaneously learning both probability function and distributed representation of words.
% What makes NLM's strength questionable and what is its consequences? 
%However, learning representations of rare words with insufficient training samples remains challenging, leading to the sub-optimal performance of NLMs.
However, learning representations of rare words remains challenging, leading to the sub-optimal performance of NLMs.
The problem escalates when a rare word is a named entity such as names of persons or locations, which are important keywords for many downstream tasks.
% How existing works address the problem?
Currently, a common practice is to exclude the rare words from the vocabulary by treating them as out-of-vocabulary or to ignore the rare word problem and train the NLM with full vocabulary as usual.
%In NLM, the common practice is to treat the rare words as out-of-vocabulary with unique representation or to break them down into finer level linguistic units.
Instead, we propose to enrich the representations of rare words in pre-trained NLM by borrowing linguistic knowledge from related words.
The proposed approach doesn't require additional training data and expensive post-training procedures.
%We applied the proposed approach to enrich representations of Singapore street names in pretrained NLM and evaluated it on n-best rescoring task for the speech recognition system where consistent perplexity and word error rate improvements are achieved.
The enriched NLM was evaluated on the $n$-best rescoring task for Singapore English speech recognition containing street names as a case study.
\fi
\end{abstract}

\section{Introduction}
\label{sec:intro}

%The neural language models (NLM) have achieved a great success in many speech and language processing tasks~\cite{bengio2003neural,mikolov2010recurrent,sundermeyer2012lstm}.
%Different from the traditional count-based N-gram models that suffer from the data sparsity problem~\cite{chen1999empirical},
%the NLMs can better generalize to unseen word sequences.
%The generalization is obtained by learning the distributed vector representations of words and then expressing the probability function in terms of these vector representations~\cite{bengio2003neural}.
%As a result, the learned vector representations will reflect different aspects and commonalities of the words, and can be used to model various combination of words unseen during the training stage.

The neural language models (NLM) have achieved great success in many speech and language processing applications~\cite{bengio2003neural,mikolov2010recurrent,sundermeyer2012lstm}.
Particularly, it is highly employed in automatic speech recognition (ASR) systems to rescore the $n$-best hypotheses list where the state-of-the-art results are attained.
%In speech recognition pipeline, computationally expensive NLMs are commonly used to rescore the n-best lists or lattices generated at the decoding stage by simpler count-based N-gram models.
Different from the traditional count-based $N$-gram models that suffer from the data sparsity problem~\cite{chen1999empirical},
%the NLMs can effectively model unseen word sequences by employing continuous vector representations of words learnt during the training stage. 
the NLMs possess superior generalization capability.
The generalization is mainly achieved by learning dense vector representations of words as part of the training process and using them to express the probability function~\cite{bengio2003neural}.
%, and then expressing the probability function in terms of these vectors~\cite{bengio2003neural}.
As a result, the learned word representations capture differences and commonalities between words, and thus enable NLMs to model different combination of words including the ones unseen during the training.

However, this concept assumes that each word appears a sufficient amount of times in the training data.
%However, this concept highly depends on the availability of sufficient amount of training samples with different contexts for each word.
%in order to learn reliable word representations.
For rare words with little or no training samples, the learned representations will be poor~\cite{bahdanau2017learning}.
Consequently, the NLM will assign them unreliable probability estimates.
Moreover, the representation of the rare word will be used as a context for the neighbouring words, as such, the entire word sequence containing the rare word will be underestimated. 
The problem exacerbates when a rare word is a named entity such as names of persons, locations, organizations and so on which are important keywords for downstream tasks such as voice search~\cite{schalkwyk2010your}.
% (e.g. domain-specific terms, names of people, places and so on)
%The commonly accepted approach in NLMs is to treat the rare words as out-of-vocabulary words by mapping them to special $<$unk$>$ token~\cite{park2010improved} or .

Currently, a common practice in language modeling is to ignore the rare word problem.
For example, by limiting the NLM's vocabulary set to the most frequent words and treat the remaining words as out-of-vocabulary (OOV), i.e. mapping them to special $<$\textit{unk}$>$ token~\cite{park2010improved} or train the NLM with full vocabulary as usual.
%Currently, a common practice is to limit the NLM's vocabulary set to the most frequent words and treat the remaining words as out-of-vocabulary (OOV), i.e. mapping them to special $<$\textit{unk}$>$ token~\cite{park2010improved}.
%Another practice is to ignore the rare word problem and train the model with full vocabulary as usual.
%Another practice is to train the model with full vocabulary as usual.
%Alternatively, the rare words can be also broken down into finer level linguistic units such as characters~\cite{kim2016character}, syllables~\cite{mikolov2012subword} or morphemes~\cite{qiu2014co}.
%In NLMs, the temporal remedy to the rare word problem is to treat them as out-of-vocabulary words, i.e. mapping them to special $<$unk$>$ token~\cite{park2010improved}, or to break them down into finer level linguistic units such as characters~\cite{kim2016character} or syllables~\cite{mikolov2012subword}.
The former approach conflates all the meanings of rare words into a single representation, losing the properties of individual words.
The latter approach will result in low-quality rare word representations.
For both approaches, the probability estimates of hypotheses incorporating the rare words will be unreliable, leading to the sub-optimal performance of NLM.
%On the other hand, the later approach successes at capturing properties of morphologically related words (e.g. "run" vs "running"),
%but may fail to capture distinctions between semantically unrelated words (e.g. "run" vs "rung")~\cite{bahdanau2017learning,bojanowski2016enriching}.

In this work, we propose an efficient method to enrich the vector representations of rare words in pre-trained NLMs.
The proposed method augments the word embedding matrices of pre-trained NLM while keeping other parameters unchanged.
Specifically, our method shifts the rare word representation towards its semantic landmark in the embedding space using representations of other semantically and syntactically similar words.
%Specifically, our method updates the embedding vectors of rare words using embedding vectors of other semantically and syn-tactically similar words. %The proposed method is based on embedding matrices augmentation technique where embedding vectors of rare words are updated using embedding vectors of other semantically and syntactically similar words.
%Specifically, our method updates the embedding vectors of rare words using embedding vectors of other semantically and syntactically similar words. 
%Our method is inspired by~\cite{pilehvar2016conflated}, where representations of rare words are induced using the representations of semantically and syntactically similar words.
%Specifically, our method shifts the rare word representation towards its semantic landmark in the embedding space.
%Our method is inspired by~\cite{pilehvar2016conflated}, where rare word representations are induced using representations of semantically and syntactically similar words.
%Specifically, the method places the rare word in the region of the semantic space in the proximity of its semantic landmarks.
This method has been shown effective for word similarity task~\cite{pilehvar2017inducing} and vocabulary expansion for NLMs~\cite{khassanov2018unsupervised}.
We further extend its application to the rare word representation enrichment.
%This paper further extends its application to enrich the representations of rare words in NLMs.
%and apply it to rescore the n-best lists from the state-of-the-art Singapore English speech recognition task.
To evaluate the proposed method, we first enrich the representations of rare Singapore street names in pre-trained NLM and then use it to rescore the $100$-best hypotheses output from the state-of-the-art Singapore English ASR system.
The enriched NLM reduces the word error rate by $6\%$ relative and improves the recognition accuracy of the rare words by $16\%$ absolute as compared to the strong baseline NLM.

\ifx
To achieve this, we exploit the structure of NLMs which can be decoupled into three main parts: 1) input projection layer, 2) middle layers, and
3) output projection layer as shown in Figure~\ref{fig:nlm}.
The input and output layers are parameterized by word embedding matrices used to transform words from one form to another, whereas the parameters of the middle layers are used to yield high-level context features.
%The roles of the input and output layers is to map vectors from high to low and low to high dimensions respectively, whereas middle layers are used to learn high-level features.
%The role of the input layer is to map one-hot encodings of the words to the low-dimensional continuous vector space, whereas the middle layers are used to learn high-level context features.
%Lastly, the output layer, also called softmax weights, projects the context vector into higher dimensional space to compute the scores of the next potential words to follow given context.
The proposed method updates the embeddings of rare words in the input and output projection layers while keeping the parameters of the middle layers unchanged.
Our method keeps the functionality of pre-trained NLM intact as long as rare words are properly updated.
%Moreover, proposed approach doesn't require additional data as well as re-training and post-training procedures.
To evaluate the proposed method, we first enrich the representation of rare Singapore street names in pre-trained NLM and then use the enriched NLM to rescore the $n$-best hypotheses output from the state-of-the-art Singapore English ASR system.
As a result, we expect the underestimated hypotheses incorporating the rare street names to ascend to the top of the list.
%We applied this method to enrich the representation of Singapore street names in pre-trained NLM.
%The enriched NLM was then evaluated on the $n$-best list rescoring task from the state-of-the-art Singapore English speech recognizer where consistent  word error rate (WER) improvements are achieved.
%We evaluated the enriched NLM on the n-best list rescoring task from the state-of-the-art Singapore English speech recognizer where consistent perplexity and word error rare improvements are achieved over the strong baselines.
\fi

The rest of the paper is organized as follows. 
%Section 2 first reviews related works on the rare word representation induction followed by the review of approaches designed to deal with the rare word problem in language models.
Section 2 reviews related approaches designed to deal with the rare word problem.
In Section 3, we briefly describe the architecture of baseline NLM.
Section 4 presents the proposed embedding matrix augmentation technique.
In Section 5, we explain the experiment setup and discuss the obtained results.
Lastly, Section 6 concludes the paper.

\section{Related works}
\label{sec:related}
The continuous vector representations of words are typically derived from large unlabeled corpora using co-occurrence statistics~\cite{mikolov2013efficient,pennington2014glove}.
They became the dominant feature for many natural language processing applications achieving the state-of-the-art results.
%They are usually first pre-trained using the special models such as \texttt{skipgram}, \texttt{cbow}~\cite{mikolov2013efficient} or \texttt{glove}~\cite{pennington2014glove} and then employed in the target task.
%Similarly, they can be also directly trained on the end tasks as in NLMs.
%In this section, we first briefly review the methods designed to induce the representation of rare words in special models followed by the review of techniques proposed to deal with the rare word problem in language models.
%In this section, we first briefly review the methods derived to deal with the rare words in special models followed by the review of techniques employed in NLMs. 
%\subsection{Inducing rare word representations}
%Learning continuous representations of words has a long history in natural language processing.
%
%The word representations are typically derived from large unlabeled corpora using special models that utilize co-occurrence statistics.
%They became core components of many natural language processing tasks.
To generalize well, however, these tasks require many occurrences of each word and fall short if a word appears only a handful of times~\cite{luong2013better}.
Several approaches have been proposed to deal with the rare word problem and most of them can be classified under one of the three main categories shown below.
%These approaches can be classified into three main branches explained next.
%1) Breaking down words into finer level

\textbf{1) Morphological word representations.} 
%induce rare word representations.
%Another direction of works leverage finer level linguistic information by breaking down words into subword units.
A bunch of proposed works resorts to subword level linguistic units by breaking down the words into morphemes~\cite{luong2013better,lazaridou2013compositional,qiu2014co}.
%to deal with the rare word problem.
For example,~\cite{luong2013better} represented words as a function of its morphemes where the recursive neural network is applied over morpheme embeddings to obtain the embedding for a whole word.
%In~\cite{lazaridou2013compositional}, authors adapt various composition methods derived for larger chunks of text such as phrases or sentences to the morphological setting.
%A hybrid word-morpheme \texttt{cbow} model was proposed in~\cite{qiu2014co}, where word and morpheme embeddings are co-learned.
While such works have been proven effective to deal with the infrequent word variations, they depend on the morphological analyzer such as \texttt{Morfessor}~\cite{creutz2007unsupervised} and unable to model words whose morphemes are unseen during the training stage.

%While such models have proved useful, they require morphological tagging as a preprocessing step.

\textbf{2) Character-level representations.} To alleviate the rare word problem, finer level linguistic units such as syllables and characters have been also studied~\cite{mikolov2012subword,ling2015finding,kim2016character,bojanowski2016enriching}.
For example,~\cite{mikolov2012subword} explored both word-syllable and word-character level hybrid NLMs where most frequent words are kept unchanged, while rare words are split into the syllables and characters, respectively.
In similar fashion,~\cite{ling2015finding} and~\cite{kim2016character} examined character-aware NLM architectures which rely only on character level inputs, but predictions are still made at the word level.
%The former approach employs convolutional neural networks over characters, while later one uses bi-directional long short-term memory (LSTM) networks.
%Decomposition of words into character $n$-grams is investigated in~\cite{bojanowski2016enriching} where word embeddings are obtained by summing corresponding $n$-gram vectors.

Character-level models eliminate the need for morphological tagging or manual feature engineering and they comprise substantially fewer number of parameters compared to word-level models.
Moreover, these approaches success at capturing properties of morphologically related words (e.g. `run' vs `running'),
but may fail to capture distinctions between semantically unrelated words (e.g. `run' vs `rung')~\cite{bojanowski2016enriching,bahdanau2017learning}.
%Nevertheless, character-level models are generally outperformed by word-level models~\cite{mikolov2012subword}.

%The advantage of these models is that they have substantially fewer number of parameters to learn compared to standard word level models.
%Although these models have substantially fewer number of parameters to learn, they are usually outperformed by standard word based models~\cite{mikolov2012subword}.
%The advantage of these models is that they have substantially lower number of parameters to learn.
%In~\cite{bojanowski2016enriching}, authors proposed to decompose words into bag of character $n$-grams, the word embeddings are obtained by summing these $n$-gram vectors.

\ifx
\textbf{3) Mixture of features.} While aforementioned approaches use only surface form to represent words and subwords, another group of works focus on NLM architectures that utilize a set of different features jointly.
Particularly,~\cite{alexandrescu2006factored} proposed factored NLM where in addition to the surface form the auxiliary feature vectors are first concatenated and then input to the model.
Similar approach is proposed in~\cite{botha2014compositional} where various feature type vectors are summed instead\footnote{In this paper, authors used only morpheme features.}.
Both models are able to handle rare words by resorting to more general feature types.
%The features are usually obtained by factorizing words into its morpheme components, lemma, part of speech and so on.
The commonly used features include morphemes, lemma, part of speech tags and so on.
\fi

\textbf{3) Knowledge-powered word representations.} Another direction of works leverage external knowledge to enhance representations of rare words~\cite{bahdanau2017learning,xu2014rc,faruqui2014retrofitting}.
%2) Using word definitions: "Bahdanau, Dzmitry, et al. "Learning to compute word embeddings on the fly."
For example,~\cite{bahdanau2017learning} employed word definitions obtained from the \texttt{WordNet}~\cite{miller1995wordnet} to model rare words on the separate network.
%2) Using context: "A La Carte Embedding: Cheap but Effective Induction of Semantic Feature Vectors"
%Similar approach has been proposed in~\cite{khodak2018carte} where linear transformation of context words around the rare word is used instead.
%Another interesting approach is proposed by~\cite{xu2014rc} where external knowledge is incorporated as a regularization term to \texttt{skip-gram} model's objective function.
Alternatively,~\cite{xu2014rc} proposed to incorporate external knowledge as a regularization term to the original model's objective function.
Although these approaches have shown promising results, they highly depend on the availability of external hand engineered lexical resources.

Note that the aforementioned approaches can be also used jointly.
For example, by using factored NLM architecture~\cite{alexandrescu2006factored} where different feature types can be combined.

%Different approach to deal with rare words has been proposed in~\cite{pilehvar2017inducing} where representation of rare words are induced by borrowing knowledge from the similar words.
%This approach has been proven effective for special models designed to learn word embeddings such as \texttt{skipgram}, \texttt{cbow}~\cite{mikolov2013efficient} or \texttt{glove}~\cite{pennington2014glove}.
%We further extend it to enrich representation of rare named-entities in NLMs for both input and output layers.
%In the next section, we will present a methodology that enables to apply this method in the context of NLMs.

%We first briefly describe the architecture of NLM and define used notations to facilitate understanding of the proposed methodology in the later part.

\section{Baseline NLM architecture}
The NLM architectures can be generally classified into two main categories: feedforward~\cite{bengio2003neural} and recurrent~\cite{mikolov2010recurrent}.
Our method can be applied to both of them, but in this paper we will focus on recurrent architecture with LSTM units which has been shown to achieve the state-of-the-art results~\cite{sundermeyer2013comparison}.
%Different NLM architectures have been proposed in recent years such as feedforward~\cite{bengio2003neural}, recurrent~\cite{mikolov2010recurrent} and long short-term memory (LSTM)~\cite{sundermeyer2012lstm}.
%Even though our method can be applied to all mentioned NLM architectures, in this paper, .

\begin{figure}[t]
    \centering
    \includegraphics[width=\linewidth]{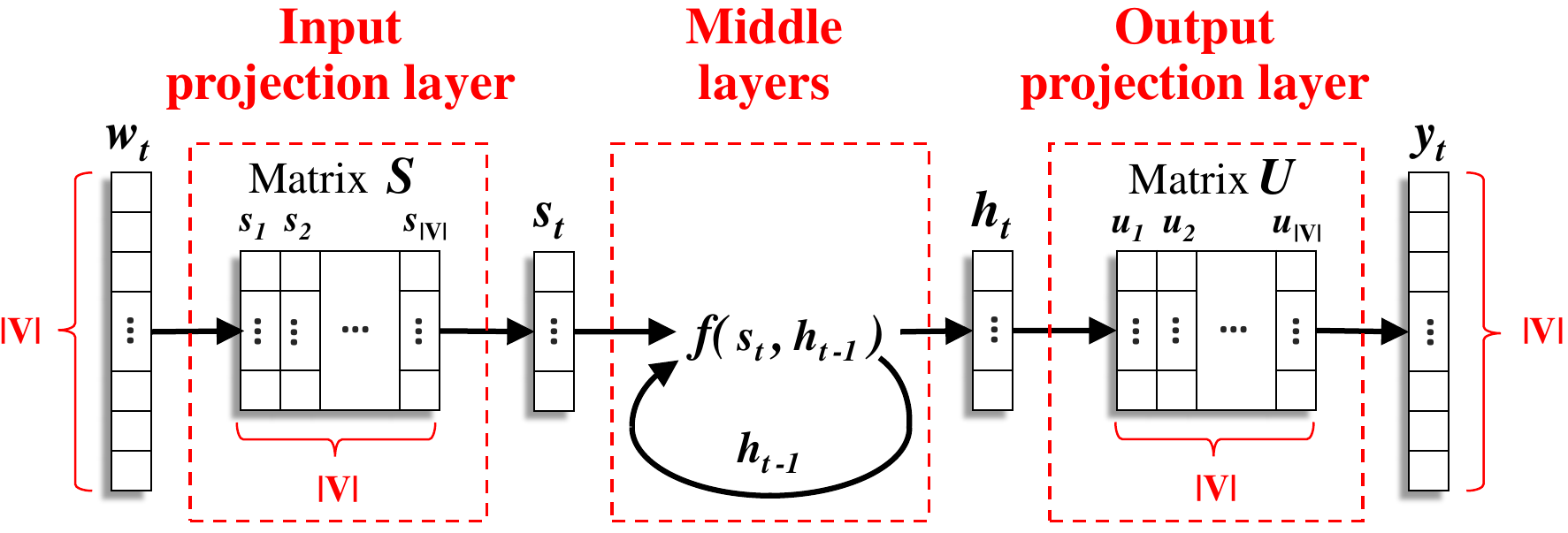}
    \caption{NLM architecture decomposed into three components.}
    \label{fig:nlm}
\end{figure}

The conventional recurrent LSTM architecture can be decoupled into three main components as shown in Figure~\ref{fig:nlm}: 1) input projection layer, 2) middle layers, and 3) output projection layer.
The input layer is parameterized by input embedding matrix $\mathbf{S}$ used to map one-hot encoding representation of word $w_t\in\mathbb{R}^{|V|}$ at time $t$ into continuous vector representation $s_t$, where $|V|$ is a vocabulary size:
\begin{equation}
    s_t=\mathbf{S}w_t
\end{equation}

The embedding vector $s_t$ and a high-level context feature vector from the previous time step $h_{t-1}$ are then combined by non-linear middle layers, which can be represented as function $f()$, to produce a new context feature vector $h_t$:
\begin{equation}
    h_t=f(s_t,h_{t-1})
\end{equation}
The non-linear function $f()$ can employ simple activation units such as ReLU and hyperbolic tangent or more complex units such as LSTM and GRU.
The middle layers can be also formed by composing several such functions.

Lastly, the context vector $h_t$ is fed to the output layer which is parameterized by output embedding matrix $\mathbf{U}$ to produce a high-dimensional vector $y_t\in\mathbb{R}^{|V|}$:
\begin{equation}
    y_t=\mathbf{U}^Th_t
\end{equation}
The entries of output vector $y_t$ represent the scores of words to follow the context $h_t$.
These scores are then normalized by softmax function to form a probability distribution.

Our method modifies the embedding matrices $\mathbf{S}$ and $\mathbf{U}$ while keeping the middle layer $f()$ intact as will be explained in the next section.

\ifx
The input layer is parameterized by input embedding matrix $\mathbf{S}\in\mathbb{R}^{d_s\times|V|}$ used to map one-hot encoding representation of word $w_t\in\mathbb{R}^{|V|}$ at time $t$ into continuous vector representation $s_t\in\mathbb{R}^{d_s}$, where $d_s$ and $|V|$ are input word embedding dimension and vocabulary size, respectively:
\begin{equation}
    s_t=\mathbf{S}w_t
\end{equation}

The embedding vector $s_t$ and a high-level context feature vector from the previous time step $h_{t-1}\in\mathbb{R}^{d_h}$ are then combined by non-linear middle layers, which can be represented as function $f()$, to produce a new high-level context feature vector $h_t\in\mathbb{R}^{d_h}$, where $d_h$ is context vector dimension:
\begin{equation}
    h_t=f(s_t,h_{t-1})
\end{equation}
The function $f()$ can be simple activation units such as ReLU and hyperbolic tangent or more complex units such as LSTM and GRU.
The middle layers can be also formed by stacking several such functions.

Lastly, the context vector $h_t$ is fed to the output layer which is parameterized by output embedding matrix $\mathbf{U}\in\mathbb{R}^{d_h\times |V|}$ to produce a high-dimensional vector $y_t\in\mathbb{R}^{|V|}$:
\begin{equation}
    y_t=\mathbf{U}^Th_t
\end{equation}
The entries of output vector $y_t$ represent the scores of words to follow the context $h_t$.
These scores are then normalized by softmax function to form a probability distribution.
\fi

\section{Embedding matrix augmentation}
We start with the assumption that we are given a pre-trained NLM which models full vocabulary including both frequent and rare words.
In such models, the rare words will be poorly represented, leading to the sub-optimal performance.
Therefore, our goal is to enrich the representation of rare words without collecting additional training data and incurring expensive post-processing procedures.
%Therefore, our goal is to enrich the representation of the rare words and consequently improve the probability estimation performance of the NLM.
%Specifically, we will place the rare word representation in the proximity of its semantic landmark using the representations of the semantically and syntactically similar words.

To achieve this goal, we exploit the structure of NLM where input and output layers are parameterized by word embedding matrices (see Figure~\ref{fig:nlm}).
Particularly, we propose to modify both input and output embedding vectors of the rare words, while keeping the parameters of middle layers unchanged.
The embedding vectors of the rare words are modified using embedding vectors of other semantically and syntactically similar words.
This approach will retain the linguistic regularities encapsulated within original pre-trained NLM, given that embeddings of the rare words are properly modified.
%Additionally, the proposed approach obviates the need for collection of expensive in-domain training data and doesn't require time-consuming post-training procedures.
Our method can be also viewed as a language model adaptation task~\cite{khassanov2017unsupervised} where instead of topic or speaking style the vocabulary is adapted to conform with the words used in the target domain.

The proposed method has three main steps: 1) identifying the rare words, 2) finding similar words and 3) enriching rare word representations.

\textbf{1) Identifying the rare words.} To identify the rare words we can simply count the frequency of words in the training data and set a frequency threshold below which all words are considered rare.
This approach, however, might result in too many rare words.
%The representation of all rare words is updated next.
%This approach, however, will be expensive if the number of rare words is too large.
To reduce the computation time, we can limit the rare words to those which appear in the $n$-best hypotheses or word lattice output.
%Instead, we can consider only the rare words present in the test data to reduce the computation time.
%For example, by extracting the rare words from the speech recognition output, e.g. $n$-best hypotheses or word lattice.
%As a result, we expect the underestimated hypotheses including the rare words to ascend.
%In our experiments, we will evaluate and compare both approaches.
%While being efficient, this approach will miss unrecognized rare words.
%Therefore, in our experiments, we will evaluate and compare both approaches.

\textbf{2) Finding similar words.} Given a subset of rare words $V_{rare}\subset V$, the next step is to select a list of similar candidate words $\mathcal{C}$ for each rare word.
The selected candidates will be used to enrich representations of rare words, hence, they must be frequent and present in the vocabulary $V$ of NLM.
In addition, they should be similar to the target rare word both in semantic meaning and syntactic behavior.
Note that selecting inadequate candidates might deteriorate the performance of NLM, thus, they should be carefully inspected.

Several effective methods exist that can find appropriate candidate words.
For example, using lexical resources that contain synonyms and related words such as~\texttt{WordNet} or employing pre-trained word embeddings from~\texttt{skip-gram} or~\texttt{cbow} models~\cite{mikolov2013efficient} which can also find similar words.
In our experiments, we use lexical resource containing a list of Singapore street names where frequent street names will be used to update representations of rare street names.
%This approach can be also viewed as a vocabulary adaptation process where representations of domain-specific terms are enriched.
%For example, to boost recognition of Singapore street names or Chinese and Malay person names in Singapore English discourses.

\textbf{3) Enriching rare word representations.}
Let $s_{r}$ be an embedding vector of some rare word $w_r$ in space defined by input embedding matrix $\mathbf{S}$ and let $\mathcal{C}_{r}$ be corresponding set of similar words.
%We adopt the method of~\cite{pilehvar2016conflated} and 
We enrich the representation of $s_{r}$ using the words in $\mathcal{C}_{r}$ by the following formula:
%We obtain the new updated embedding vector $\hat{s}_r$ using the following formula:
%We update the embedding vector $s_r$ using the words in $\mathcal{C}_{r}$ by following formula:
\begin{equation}
    \hat{s}_r=\frac{s_r+\sum_{s_{c}\in\mathcal{C}_r}m_{c}s_{c}}{|\mathcal{C}_r|+1}
    \label{eq:enrich}
\end{equation}
where $\hat{s}_r$ is the enriched representation of $s_r$, $s_{c}$ is an embedding vector of similar candidate word and $m_{c}$ is a metric used to weigh candidates based on importance. %i.e. more representative candidates should be assigned higher weights.
The $m_{c}$ can be estimated using frequency counts or similarity score where most frequent or most similar candidates are given higher weights.
In our experiments, we weigh the candidates equally.

The Eq. \eqref{eq:enrich} typically shifts the embedding of a rare word towards the weighted centroid of its semantic landmark.
%The motivation is that correlated words in terms of both semantic meaning and syntactic behavior should be placed close to each other in the embedding space.
The motivation is that highly correlated words, in terms of both semantic meaning and syntactic behavior, should be close to each other in the embedding space.
We then use the same candidates and formula to update the corresponding rare word embedding $u_{r}$ in the output embedding matrix $\mathbf{U}$.
This procedure is then repeated for the remaining words in subset $V_{rare}$.

\ifx
\begin{table}[t]
    \centering
    \begin{tabular}{c|l|r|r}
        \hline
        \textbf{No.}    & \textbf{Recording Name}   & \textbf{Duration} & \textbf{Words}\\ \hline \hline
        1               & Balestier, Ang mo kio     & 9m 57s            & 1,479         \\ 
        2               & Kreta ayer, Aljunied      & 10m 25s           & 1,207         \\ 
        3               & Marsiling, Bedok          & 7m 24s            & 862           \\ 
        4               & Boon lay                  & 9m 30s            & 1,695         \\ 
        5               & Bukit batok               & 7m 41s            & 1,801         \\
        6               & Sembawang                 & 9m 3s             & 1,018         \\ \hline
        \multicolumn{2}{c|}{Total}                  & 54m               & 7,342         \\ \hline
    \end{tabular}
    \caption{Characteristics of evaluation set.}
    \label{tab:eval_set}
\end{table}
\fi

\section{Experiment}
In this section, we describe experiments conducted to evaluate the effectiveness of the proposed embedding matrix augmentation technique.
Particularly, we first enrich the rare Singapore street name representations in pre-trained NLM and then use the enriched NLM to rescore 100-best hypotheses output from the Singapore English ASR.
The ASR system is built by Kaldi~\cite{povey2011kaldi} speech recognition toolkit using Singapore English speech corpus.
%The SE data is mostly collected by transcribing Singaporean speech obtained from parliament, talk shows and student interviews.
%Thus, the main difference of SE data from GE is prevalence of uncommon local names of persons and locations, e.g. in Singapore, Lim is more common name than John.
%Yet the available SE data is insufficient to cover all named entities of Singapore, while collecting additional SE data is slow and expensive process.
To highlight the importance of enriching the rare word representations, we used 1 hour recording of 9 read articles about Singapore streets as an evaluation set\footnote{\url{https://github.com/khassanoff/SG_streets}} (7.3k words).

We compare our enriched model against three state-of-the-art language models (LM) including Kneser-Ney smoothed 4-gram\footnote{We also examined other $N$-gram models and found 4-gram to work best for our case.} (KN4), Kaldi-RNNLM~\cite{xu2018neural} and recurrent LSTM LM~\cite{sundermeyer2012lstm}.
Our model is obtained by enriching the representations of rare Singapore street names in the recurrent LSTM LM, and we call it E-LSTM.
%As the baseline language models (LM), we trained three state-of-the-art models including Kneser-Ney smoothed 4-gram (KN4)\footnote{We also examined other $N$-gram models and found 4-gram to work best for our case.}, Kaldi-RNNLM~\cite{xu2018neural} and recurrent LSTM LM.
%Our model is obtained by enriching the representations of rare Singapore street names in the baseline recurrent LSTM LM, and we call it E-LSTM.
The performance of these four LMs is evaluated on the 100-best rescoring task.

%\textbf{Experiment setup.}
\subsection{Experiment setup}
\textbf{Acoustic model.} The acoustic model (AM) is built using `nnet3+chain' setup of Kaldi and trained on 270 hours of transcribed Singapore English data which mostly consist of speech taken from parliament, talk shows and interviews.

\textbf{Lexicon.} The lexicon is constructed by assembling $51k$ unique words which include around $2k$ Singapore street names.
To avoid ambiguity, the street names consisting of more than one word are joined using the underscore symbol, e.g. `Boon lay' is changed to `Boon\_lay'.
This lexicon was also used as a vocabulary set for LMs.

\textbf{Language model.} To train LMs, we used AM transcripts and web crawled Singapore related data which resulted in total 1M in-domain sentences (16M words).
In addition, we used Google's 1 billion word (1BW) benchmark corpus~\cite{chelba2013one} to account for generic English word sequence statistics.

The KN4 is trained on combined in-domain (AM transcript$+$web crawled) and generic 1BW data. 
It was built using SRILM toolkit~\cite{stolcke2002srilm} with $51k$ vocabulary set.
We used KN4 model to rescore both word lattice and $100$-best list.
Its pruned version KN4\_pruned was used during the decoding stage.

%The Kaldi-RNNLM jointly models word and character $n$-gram features where most frequent words are kept while rare words are decomposed into character n-grams.
The Kaldi-RNNLM is a word-character level hybrid model designed to overcome the rare word problem by decomposing the rare words into character $n$-grams while keeping the most frequent words unchanged.
%Specifically, it keeps the most frequent words and decomposes the rare words into character $n$-grams.
%We train it using only in-domain data as a $2$-layer LSTM with $800$ units in each layer.
It was trained as a $2$-layer LSTM\footnote{Changing the number of layers and its size didn't improve WER.
} with $800$ units in each layer using only in-domain data.
The input and output embedding matrices were tied and embedding space dimension was set to $800$.
For vocabulary, we tried to keep a different number of most frequent words and found $20k$ to perform best, the remaining 31k words were decomposed.
%The remaining hyper-parameters were set to recommended values.

The recurrent LSTM LM is a word level model which was built using our own implementation in PyTorch~\cite{paszke2017automatic}.
It was trained as a single layer LSTM with $1k$ units using in-domain data and $51k$ vocabulary set.
The input and output embedding space dimensions were set to $300$ and $1000$, respectively.
The parameters of the model were learned by truncated BPTT~\cite{werbos1990backpropagation} and SGD with gradient clipping.
We also applied dropout for regularization~\cite{zaremba2014recurrent}.

Lastly, our E-LSTM model is obtained by enriching the rare word representations in the pre-trained recurrent LSTM LM.
As a case study, we use Singapore street names\footnote{\url{https://geographic.org/streetview/singapore/}} where frequent streets will be used to enrich the rare street representations. 
%Particularly, we employ the list of Singapore street names\footnote{https://geographic.org/streetview/singapore/} where frequent streets will be used to enrich the rare streets.
In particular, we first count the frequency of each street name in the in-domain data and then divide them into two subsets of frequent and rare streets using some threshold value.
Next, we randomly choose words from the subset of frequent streets and employ them to enrich the representations of all rare streets (allStreets) using the Eq.~\eqref{eq:enrich}.
To reduce computation time, we also tried to enrich only rare streets present in the $100$-best hypotheses output (fromNbest).
%As a result, we expect the underestimated hypotheses including the rare street names to ascend on top of the list.
%To identify the list of rare words for the update, we examined two different approaches.
%In the first approach, the representations of all rare street names are updated, we call it \textit{allStreets}.
%While in the second approach, only rare street names present in the $100$-best hypotheses output are updated, we call it \textit{fromNbest}.

\subsection{Experiment results}
The experiment results are shown in Table~\ref{tab:wer}.
In these experiments, we divide the street names into frequent and rare subsets using the threshold value of 10.
To enrich the rare streets we used 5 randomly chosen frequent street names\footnote{For consistency, we fix the chosen frequent streets to be same.}.
The initial word error rate (WER) without any rescoring is $17.07\%$.

%As shown in table~\ref{tab:wer}, the initial word error rate (WER) without any rescoring is $17.07\%$.

%We limit the number of selected frequent words to five for faster computation.
The obtained results show that the E-LSTM model outperforms the strong KN4 used to rescore the word lattice by $16\%$ relative WER (from $16.52\%$ to $13.83\%$).
Moreover, it achieves $6\%$ relative WER improvement over Kaldi-RNNLM and LSTM models (from $14.73\%$ to $13.83\%$).
We found that enriching only rare streets present in the 100-best hypotheses (\textit{fromNbest}) achieves a similar result as enriching all rare streets (\textit{allStreets}), while \textit{fromNbest} being much faster.

The state-of-the-art WER results are usually achieved by interpolating NLM and count-based $N$-gram models which have been shown to complement each other~\cite{mikolov2010recurrent,sundermeyer2012lstm}. 
To this end, we interpolated\footnote{Interpolation weight for KN4 is set to $0.3$.} NLMs with KN4 and achieved further WER reductions.
Interestingly, the baseline LSTM model doesn't benefit from KN4\footnote{Changing the interpolation weight didn't help.}, while E-LSTM gains additional $2\%$ relative WER improvement (from $13.83\%$ to $13.55\%$). 
%Further WER reductions are achieved by interpolating\footnote{Interpolation weight for KN4 is set to $0.3$.} models with KN4, except for the baseline LSTM.
%Importantly, all these improvements are achieved without using any additional in-domain data and expensive post-training procedures.

\begin{table}[t]
    \caption{The perplexity and WER results on evaluation set}
    \label{tab:wer}
    \centering
    \begin{tabular}{l|c|c|c}
        \toprule
        \textbf{LM}                 & \textbf{Perplexity}   & \textbf{Rescore}  & \textbf{WER}      \\ \hline \hline 
        KN4\_pruned                 & 436                   & -                 & 17.07\%           \\ \hline
        \multirow{2}{*}{KN4}        & \multirow{2}{*}{351}  & Lattice           & 16.52\%           \\
                                    &                       & 100-best          & 16.84\%           \\ \hline
        Kaldi-RNNLM                 & -                     & 100-best          & 14.73\%           \\ 
        \multicolumn{1}{l|}{+KN4}   & -                     & 100-best          & 14.10\%           \\ \hline
        LSTM                        & 295                   & 100-best          & 14.74\%           \\ 
        \multicolumn{1}{l|}{+KN4}   & -                     & 100-best          & 14.95\%           \\ \hline
        E-LSTM (allStreets)         & 242                   & 100-best          & 13.87\%           \\ 
        \multicolumn{1}{l|}{+KN4}   & -                     & 100-best          & 13.58\%           \\ \hline
        E-LSTM (fromNbest)          & 234                   & 100-best          & \textbf{13.83}\%  \\ 
        \multicolumn{1}{l|}{+KN4}   & -                     & 100-best          & \textbf{13.55}\%  \\
        \bottomrule
    \end{tabular}
\end{table}

\subsubsection{Changing the frequency threshold value}
\begin{figure}[h]
    \centering
    \includegraphics[width=0.95\linewidth]{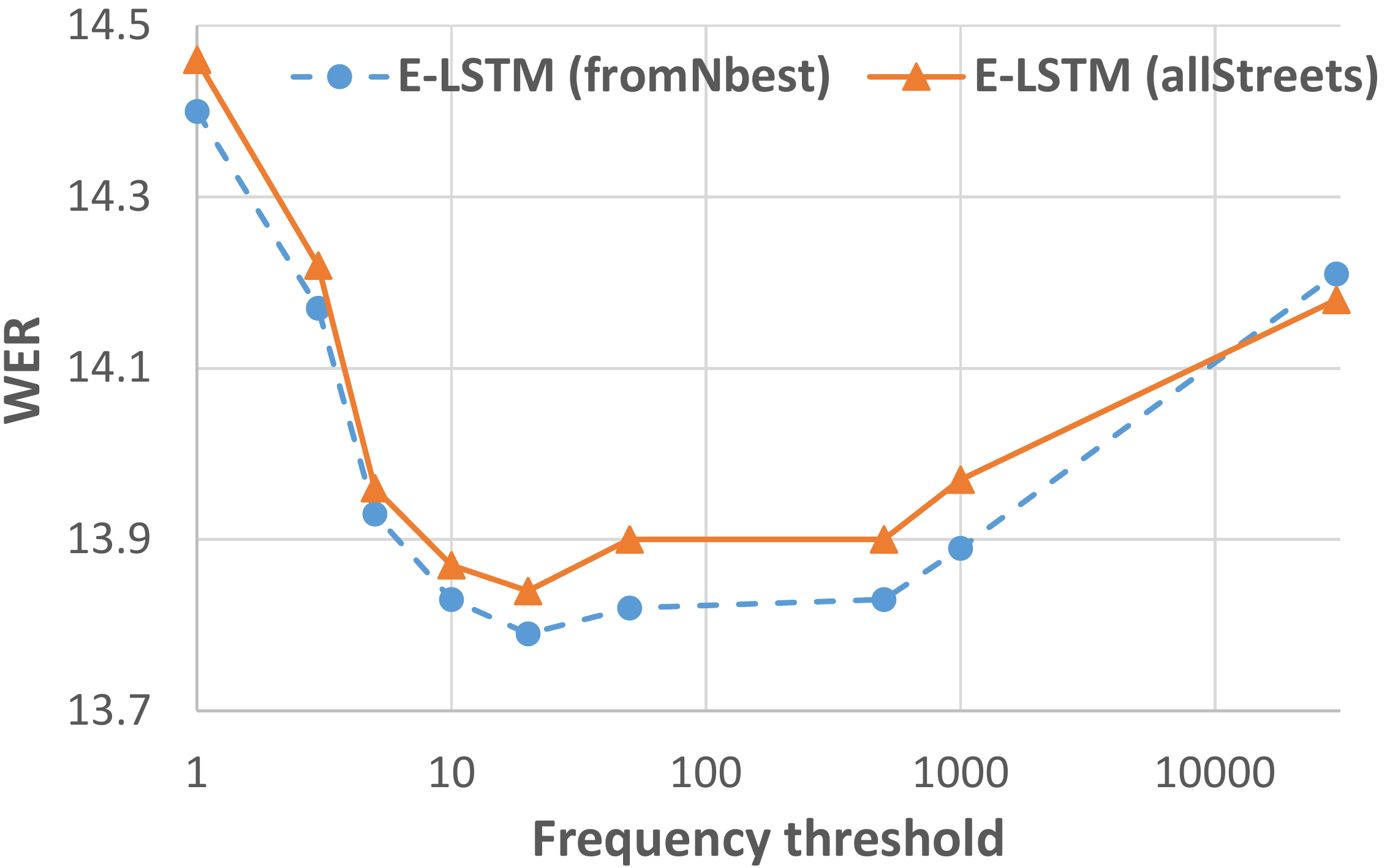}
    \caption{WER performance of E-LSTM at different frequency threshold values used to split frequent and rare street names.}
    \label{fig:th_vs_wer}
\end{figure}

To determine the effective frequency threshold range, used to split the street names into frequent and rare subsets, we repeat the experiment with different threshold values as shown in Figure~\ref{fig:th_vs_wer}.
We observe that setting it between $5$ and $50$ is sufficient to achieve good results.
On the other hand, setting it too low or high will deteriorate WER as can be seen from the left and right tails of the plot in Figure~\ref{fig:th_vs_wer}.
%The reason is because low threshold will result in very few rare words, while high threshold will cause to update word representations which already had sufficient amount of training samples.

\subsubsection{Changing the number of frequent words}
We also repeat the experiment to determine the optimal number of frequent words to use to enrich the rare words.
% we repeat the experiment with different number of $k$ frequent words where $k$ was increased from $0$ up to $100$ with step size of $10$.
%For consistency, in each experiment we select most frequent $k$ streets.
We observed that for all cases the WER results are similarly good.
For fast computation, we recommend to use around $3$-$20$ most frequent words.
This experiment is incomplete as we didn't examine the quality of selected frequent words which requires more substantial analysis.
Due to the space limitations, we leave the further analysis for future work.

\subsubsection{Recognition accuracy of enriched rare words}
To ensure that WER improvements are achieved as a result of correctly recognizing the enriched rare street names, we compute the recognition accuracy of $265$ rare street names (see Table~\ref{tab:accu}).
%These rare streets appear $265$ times in the evaluation set.
The experiment results show that after enriching the baseline recurrent LSTM LM, the recognition accuracy is increased by $16.23\%$ (from $43.02\%$ to $59.25\%$) achieving the best result among all LMs.
Furthermore, we observe that correctly recognizing the rare street names also helps to recover neighbouring words (see Table~\ref{tab:examples}).
These results confirm the effectiveness of the proposed method.

\begin{table}[t]
    \caption{Recognition accuracy of 265 rare street names}
    \label{tab:accu}
    \centering
    \begin{tabular}{l|c|c}
        \toprule
        \textbf{LM}                 & \textbf{Rescore}  & \textbf{Accuracy} \\ \hline \hline
        KN4\_pruned                 & -                 & 37.36\%           \\ 
        KN4                         & Lattice           & 36.98\%           \\
        Kaldi-RNNLM                 & 100-best          & 43.40\%           \\
        LSTM                        & 100-best          & 43.02\%           \\
        E-LSTM (fromNbest)          & 100-best          & \textbf{59.25}\%  \\
        \bottomrule
    \end{tabular}
\end{table}

\begin{table}[h]
    \footnotesize
    \caption{Examples of correctly recovered neighbouring words after rescoring with E-LSTM}
    \label{tab:examples}
    \centering
    \begin{tabular}{l|p{5.8cm}}
        \toprule
        \textbf{LM}     & \textbf{Example}  \\ \hline \hline
        \multirow{3}{*}{KN4\_pruned}& 1) a hawker centre and market began operations at \textbf{bully plays} in nineteen seventy six\\
                        & 2) by nineteen ninety four when the \textbf{book} development \textbf{guide} plan was announced \\ \hline
%                        & 2) present day \textbf{will last year} is also associated with its ubiquitous lighting shop\\\hline

        \multirow{2}{*}{LSTM}   & 1) a hawker centre and market began operation at \textbf{bully police} in nineteen seventy six \\
                        & 2) by nineteen ninety four when the \textbf{product} development \textbf{gap} plan was announced \\ \hline
%                        & 2) present day \textbf{by leisure} is also associated with its ubiquitous lighting shops \\\hline
        \multirow{2}{*}{E-LSTM} & 1) a hawker centre and market began operations at\textbf{ boon\_lay place} in nineteen seventy six\\
                                & 2) by nineteen ninety four when the \textbf{bedok} development \textbf{guide} plan was announced \\
%                                & 2) present day \textbf{balestier} is also associated with its ubiquitous lighting shops \\

        \bottomrule
    \end{tabular}
\end{table}

\section{Conclusions}
%Good word representations are important to achieve good generalization in NLMs.
In this work, we proposed an effective method to enrich the representations of rare words in pre-trained NLM.
The proposed method augments the embedding matrices of pre-trained NLM while keeping other parameters unchanged.
Importantly, it doesn't require additional in-domain data and expensive post-training procedures.
We applied our method to enrich the rare Singapore street names in pre-trained LSTM LM and used it to rescore the $100$-best list generated by the state-of-the-art Singapore English ASR system.
The enriched LSTM LM achieved $6\%$ relative WER improvement over the baseline LSTM LM.
In comparison to other strong baseline LMs, our method achieves significant WER improvements, i.e. $6\%$ and $16\%$ improvement over Kaldi-RNNLM and KN4, respectively.
Moreover, the enriched LSTM increased the recognition accuracy of rare street names by $16\%$ absolute.
We believe that the proposed method can benefit other models with similar network architecture and be easily adapted to other scenarios.

\section{Acknowledgements}
This work is supported by the project of Alibaba-NTU Singapore Joint Research Institute.

\bibliographystyle{IEEEtran}

\bibliography{mybib}

% \begin{thebibliography}{9}
% \bibitem[1]{Davis80-COP}
%   S.\ B.\ Davis and P.\ Mermelstein,
%   ``Comparison of parametric representation for monosyllabic word recognition in continuously spoken sentences,''
%   \textit{IEEE Transactions on Acoustics, Speech and Signal Processing}, vol.~28, no.~4, pp.~357--366, 1980.
% \bibitem[2]{Rabiner89-ATO}
%   L.\ R.\ Rabiner,
%   ``A tutorial on hidden Markov models and selected applications in speech recognition,''
%   \textit{Proceedings of the IEEE}, vol.~77, no.~2, pp.~257-286, 1989.
% \bibitem[3]{Hastie09-TEO}
%   T.\ Hastie, R.\ Tibshirani, and J.\ Friedman,
%   \textit{The Elements of Statistical Learning -- Data Mining, Inference, and Prediction}.
%   New York: Springer, 2009.
% \bibitem[4]{YourName17-XXX}
%   F.\ Lastname1, F.\ Lastname2, and F.\ Lastname3,
%   ``Title of your INTERSPEECH 2019 publication,''
%   in \textit{Interspeech 2019 -- 20\textsuperscript{th} Annual Conference of the International Speech Communication Association, September 15-19, Graz, Austria, Proceedings, Proceedings}, 2019, pp.~100--104.
% \end{thebibliography}

\end{document}